\documentclass[conference]{IEEEtran}

\IEEEoverridecommandlockouts

\usepackage{graphicx}
\usepackage{amsmath, amssymb}
\usepackage{epsfig}
\usepackage{enumerate}
\usepackage{url}
\usepackage{times}
\usepackage{psfrag}
\usepackage{cleveref}
\usepackage{algorithm}
\usepackage{algpseudocode}
\usepackage[format=default,footnotesize]{caption}
\usepackage{multirow}
\usepackage{cite}
\usepackage{comment}
\usepackage{booktabs}
\usepackage{subcaption}
\usepackage{soul}
\usepackage{rotating}
\usepackage{tabularray}
\usepackage{ragged2e}
\usepackage{upgreek}
\usepackage{wrapfig, lipsum}
\usepackage{color}

\crefname{equation}{}{}

\makeatletter
\let\myorg@bibitem\bibitem
\def\bibitem#1#2\par{%
  \@ifundefined{bibitem@#1}{%
    \myorg@bibitem{#1}#2\par
  }{%
    \begingroup
      \color{\csname bibitem@#1\endcsname}%
      \myorg@bibitem{#1}#2\par
    \endgroup
  }%
}
\makeatother

\makeatletter
\newcommand{\linebreakand}{%
  \end{@IEEEauthorhalign}
  \hfill\mbox{}\par
  \mbox{}\hfill\begin{@IEEEauthorhalign}
}
\makeatother

\let\OLDthebibliography\thebibliography
\renewcommand\thebibliography[1]{%
  \OLDthebibliography{#1}%
  \setlength{\parskip}{0pt}%
  \setlength{\itemsep}{0pt plus 0.3ex}%
}

\begin{document}

\title{Sequential Learning-Based Physics-Informed Neural Network Modeling for Parameter Estimation on a Power Converter}

\author{
    \IEEEauthorblockN{Samuela Rokocakau}
    \IEEEauthorblockA{
        \textit{ABB Corporate Research Center}\\
        Västerås, Sweden \\
        Email: samuela.rokocakau@abb.se.com
    }
    \and
    \IEEEauthorblockN{Panagiotis Kakosimos}
    \IEEEauthorblockA{
        \textit{ABB Corporate Research Center}\\
        Västerås, Sweden \\
        Email: panagiotis.kakosimos@se.abb.com
    }
    \linebreakand
        \IEEEauthorblockN{Giulia Tresca}
    \IEEEauthorblockA{
        \textit{Dept. of Electrical, Computer and Biomedical Engineering}\\
        \textit{University of Pavia}, Pavia, Italy \\
        Email: giulia.tresca@unipv.it}
        \and
    \IEEEauthorblockN{Pericle Zanchetta}
    \IEEEauthorblockA{
        \textit{Dept. of Electrical, Computer and Biomedical Engineering}\\
        \textit{University of Pavia}, Pavia, Italy \\
        Email: pericle.zanchetta@unipv.it}
}

\maketitle

\begin{abstract}\label{sect:abs}
A staged physics-informed neural network (PINN) is proposed for identifying the inductance, capacitance and parasitic resistances of a DC buck converter from time-domain data.  
Gradients are released in the order of \(L \rightarrow C \rightarrow r_{L}\rightarrow r_{C}\), a sequence derived from sensitivity analysis.  
The five-stage schedule recovers \(L\) and \(C\) to within 5\% and the series ESR to 10\%, while a three-stage variant attains the same accuracy 11\% faster.  
An \(8^{4}\) factorial sweep confirms that the trained PINN remains accurate for simultaneous \(\pm8\,\%\) perturbations of all four components, achieving a waveform RMSE of \(2.6\times10^{-2}\) over 5 ms—an order-of-magnitude better than a simultaneous-training baseline.\\

\end{abstract}

\begin{IEEEkeywords}
Physics Informed Neural Network, fault diagnostics, Buck Converter, neural network, power converter, parameter estimation, parasitic resistance, condition monitoring
\end{IEEEkeywords}

\vspace{15pt}
\section{Introduction} \label{sec:Intro}\vspace{2pt} 

\IEEEPARstart{T}{he} increasing use of electrical component parameter estimation is gaining traction across power electronics, electric vehicles, smart grids, and mission-critical systems. In such systems, accurate knowledge of component values and their parasitic elements is essential for condition monitoring, fault detection, predictive maintenance, and high-performance control \cite{Farzad_2024, Williamson_2015, Yang_2010, Wu_2025, Wen_2024}. Power converters, in particular, rely on accurate modeling to minimize downtime, reduce financial losses, and improve operational safety and efficiency.

Conventional system identification methods fall broadly into parametric and non-parametric categories. Non-parametric techniques often assume steady-state operation and use frequency domain measurements, requiring complex setups and offline post-processing \cite{Ahmeid_2017}. These methods also require intrusive excitation signals that may lead to system instability.

In contrast, parametric identification approaches — especially data-driven ones — offer flexibility by learning system dynamics from time-domain input–output data. These methods can operate during normal converter operation and avoid the need for frequency-response acquisition \cite{Zhao_2021,Gao_2023, Frede_2021}.

Recent work has emphasized the importance of identifying multiple parameters simultaneously, especially in real-world scenarios where idealized component assumptions no longer hold \cite{She_2024}. The rise of digital twins, state observers, and neural models illustrates a broader trend of using real-time, embedded estimators for live converter health and performance assessment.

Physics-Informed Neural Networks (PINNs) combine the data-driven learning power of neural networks with embedded physical constraints derived from differential equations. Unlike black-box models, PINNs integrate system-level knowledge to enforce physically consistent outputs. This makes them attractive for learning hidden dynamics and parameters of systems governed by ordinary differential equations, such as power converters\cite{Raissi_2019,Gao_2023, Yang_2025}.

Several studies \cite{Zhao_2022, She_2024, Xiang_2025} have successfully applied PINNs for estimating circuit parameters in switched-mode power supplies. These include efforts to model unknown topological durations, parasitic behaviors, and transient response—all while maintaining interpretability and structure imposed by physics.

Despite their promise, many PINN approaches in power electronics rely on complex solvers like Runge–Kutta, require full-state observability, or lack modularity for staged training. Moreover, few works isolate and study how staged parameter unfreezing affects convergence, robustness, and generalization to unseen component values.

This work presents a simplified and interpretable sequential PINN workflow tailored to a DC buck converter. It estimates four physical parameters—inductance, capacitance, inductor ESR, and capacitor parallel resistance—using staged learning. More specifically, we propose a modular training strategy for PINNs, employing both five-stage and three-stage sequential schemes that progressively activate trainable parameters to enhance convergence and interpretability. Our modeling framework incorporates parasitic resistance effects in both the inductor and capacitor branches, enriching the fidelity of the governing ODE system. To assess robustness, we conduct synthetic parameter sweeps and introduce noise-injected test cases, analyzing RMSE trends and residual compensation behavior. We further compare full and staged training strategies, highlighting trade-offs in convergence speed, accuracy, and generalizability. Finally, we present a suite of quantitative and qualitative tools for evaluating PINN performance, including RMSE heatmaps, trajectory comparisons, and physics residual analyses.

\vspace{3 pt}\section{System Overview} \label{sec:overview}\vspace{5pt}

\begin{figure}[tbp]
    \centering
    \includegraphics[width=0.9\columnwidth]{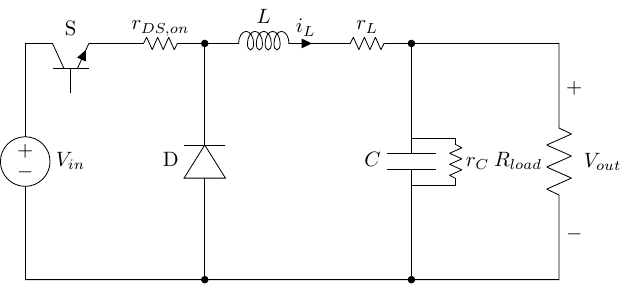}
    \caption{Buck converter with series inductor ESR \((r_L)\), switch on-resistance \((r_{DS,\text{on}})\),and capacitor leakage \((r_C)\) explicitly modeled }
    \label{fig:buck_schematic}
\end{figure}

Figure~\ref{fig:buck_schematic} depicts a non-ideal buck converter used as the plant for identification. A high-side MOSFET switch \(S\) modulates the input source \(V_{in}\) with duty ratio \(S(t)\in\{0,1\}\), and its channel on-resistance is represented by \(r_{DS,\text{on}}\).
 When \(S\) is open, the free-wheel diode \(D\) allows the inductor current~\(i_L\) to return to ground.  Energy is stored in the series inductor \(L\) and shunt capacitor \(C\), whose parasitic resistances are explicitly modeled: \(r_L\) represents copper loss in the coil, while \(r_C\) represents dielectric leakage across the capacitor plates.
 The load is a purely resistive element ($R_{load}$), and the output voltage ($V_{out}$) is measured between the positive rail and ground.  Adding \(r_{DS,\text{on}},\,r_L,\text{ and }r_C\) transforms the model from an ideal, loss-less prototype to a realistic power-stage description capable of predicting efficiency, wave-shape distortion, and damping effects observed in hardware.

The buck converter is written in continuous-time state–space form:
\begin{subequations}
\label{eq:ss_generic}
\begin{align}
  \dot{\boldsymbol x}(t) &=
  \boldsymbol f\!\bigl(\boldsymbol x(t),\,u(t),\,\boldsymbol\theta\bigr),
  \\[2pt]
  \boldsymbol x(t) &=
  \begin{bmatrix}
    \dot\imath_L(t) \\[2pt] v_O(t)
  \end{bmatrix},
  \qquad
  u(t)=S(t)
\end{align}
\end{subequations}
where \(\dot\imath_L\) is the inductor current, \(v_O\) the output-capacitor voltage and \(S(t)\in\{0,1\}\) the duty ratio of the power switch.

The parameter vector \(\boldsymbol\theta=\{L,C,r_L,r_C,R_{\text{load}},V_{\text{in}}\}\) collects the inductance, capacitance, series coil resistance, parallel leakage resistance, load resistance and input voltage, respectively.
Applying Kirchhoff’s laws under the usual assumptions of an ideal switch and negligible switch on-resistance yields the bilinear dynamics
\begin{equation}
\begin{aligned}
  \dot{\imath}_L &=
    \frac{S V_{\text{in}}-v_O - r_L\,\imath_L}{L},\\[4pt]
  \dot{v}_O &=
    \frac{\imath_L - v_O/R_{\text{load}} - v_O/r_C}{C}.
\end{aligned}
\label{eq:ss_buck}
\end{equation}


Equations ~\eqref{eq:ss_generic}-\eqref{eq:ss_buck} describe the hard physics included in the latter phases of the sequential physics-informed neural network (PINN).
During Stage 1, the neural network approximates the state derivatives on its own, allowing it to absorb measurement noise and unmodeled effects.
The PINN recovers physical parameters \(\boldsymbol\theta\) while maintaining the interpretability of a standard state-space representation by imposing \eqref{eq:ss_buck} as a differentiable constraint in Stage 2 and beyond.

\begin{figure}[tbp]
    \centering
    \includegraphics[width=0.83\linewidth]{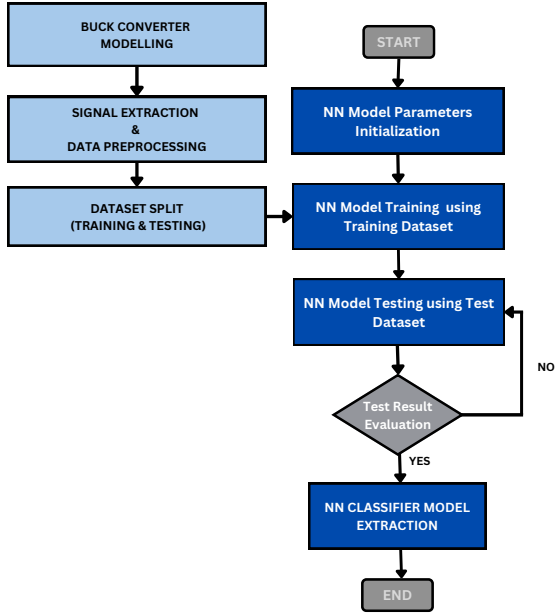}
    \caption{PINN modeling workflow.}
    \label{fig:PINN_workflow}
\end{figure}

The approach utilized in this work to construct the PINN model is a sequential approach where each stage of the training process allows for better visibility and control over the process.  Given in Fig.\ref{fig:PINN_workflow} demonstrate the workflow for the neural network modeling
In PINNs, the loss functions used is related to the residual between the neural network predictions and the true vales as defined by the mean square error. This function is defined as follows:

\paragraph*{Stage 1 — Data–only Loss}
\begin{equation}
  \mathcal L_{\mathrm d} \;=\;
  \frac{1}{N}\sum_{n=1}^{N}
  \Bigl[
    w_i\bigl(\hat{\dot{\imath}}_{L,s}^{(n)}-\dot{\imath}_{L,s}^{(n)}\bigr)^2 \;+\;
    w_v\bigl(\hat{\dot{v}}_{O,s}^{(n)}-\dot{v}_{O,s}^{(n)}\bigr)^2
  \Bigr],
  \label{eq:loss_data}
\end{equation}
where hats denote network predictions, the subscript~\(s\) indicates variables scaled to zero mean and unit variance, \(w_i,w_v\) are balancing weights, and \(N\) is the mini-batch size.

\paragraph*{Stage 2  — Combined Data \& Physics Loss}
\begin{subequations}
\begin{align}
  \mathcal L_{\mathrm{tot}}
    &= \mathcal L_{\mathrm d} + \lambda_{\mathrm{phys}}\,
       \mathcal L_{\mathrm p},
       \label{eq:loss_total}\\[2pt]
  \mathcal L_{\mathrm p}
    &= \frac{1}{N}\sum_{n=1}^{N}
       \Bigl[
         w_i^{\mathrm p}\,r_{i,s}^{2(n)} +
         w_v^{\mathrm p}\,r_{v,s}^{2(n)}
       \Bigr],
       \label{eq:loss_phys}
\end{align}
\end{subequations}
\begin{align}
  f_i &= \frac{S V_{\text{in}}-v_O-r_L \dot\imath_L}{L}, &
  f_v &= \frac{\dot\imath_L-v_O/R_{\text{load}}-v_O/r_C}{C}.
  \label{eq:rhs_analytic}
\end{align}

with scaled residuals

\begin{equation}
  r_{i,s}=\frac{\hat{\dot{\imath}}_{L}-f_i}{\sigma_{\dot{\imath}_{L}}},
  \qquad
  r_{v,s}=\frac{\hat{{v}}_{O}-f_v}{\sigma_{{v}_{O}}},
  \label{eq:scaled_residuals}
\end{equation} the empirical standard deviations used for non-dimensionalization.

In~\eqref{eq:rhs_analytic}, the symbols \(f_i\) and \(f_v\) denote the \emph{analytic right-hand sides} of the state-space equations for the buck converter, i.e. the deterministic expressions that give the ideal time derivatives of the inductor current and the capacitor voltage for a given state \(( \dot\imath_L, v_O)\), duty ratio \(S\), and parameter set \(\{L,C,r_L,r_C,R_{\text{load}},V_{\text{in}}\}\):

\begin{itemize}
\item\(f_i\) is the theoretical value of \(\dot{\imath}_L\), obtained
      from Kirchhoff’s voltage law around the power stage.
\item\(f_v\) is the corresponding theoretical value of \(\dot{v}_O\),
      derived from the capacitor-current equation.
\end{itemize}

These expressions serve as \textbf{physics targets} in the loss function: the PINN’s derivative predictions are driven to match \(\hat{\dot{\imath}}_L \approx f_i\) and \(\hat{\dot{v}}_O \approx f_v\) once the physics term is activated. The hyper-parameter \(\lambda_{\mathrm{phys}}\) weights the physics term against the data term. In the construction of the PINN model explored in this study, the data-driven and physics-driven components of the model are slightly adjusted, with parametric estimation for the converter used from a monitoring perspective using a sequential training process.

Table \ref{tab:scenarios} shows the models used in this work to contextualize the technique. The models use the PINN architecture's sequential training approach, beginning by optimizing the data-driven component of the overall to enable generalized learning system behavior. The initial stage of training focuses on the data-loss consideration of the overall loss function while determining the ODE of the overall buck converter for the specific scenario. This consideration is achieved by "freezing" the learning gradients of the observed parameters on the PDE. It is worth mentioning that for the scenarios for 2 stage training, singular parasitic resistance estimation, respectively the results were maintained good estimation performance but remained simplistic preliminary configuration leading to the other scenarios mentioned in this work.

\begin{figure*}[tbp]
    \centering
    \includegraphics[width=0.8\textwidth]{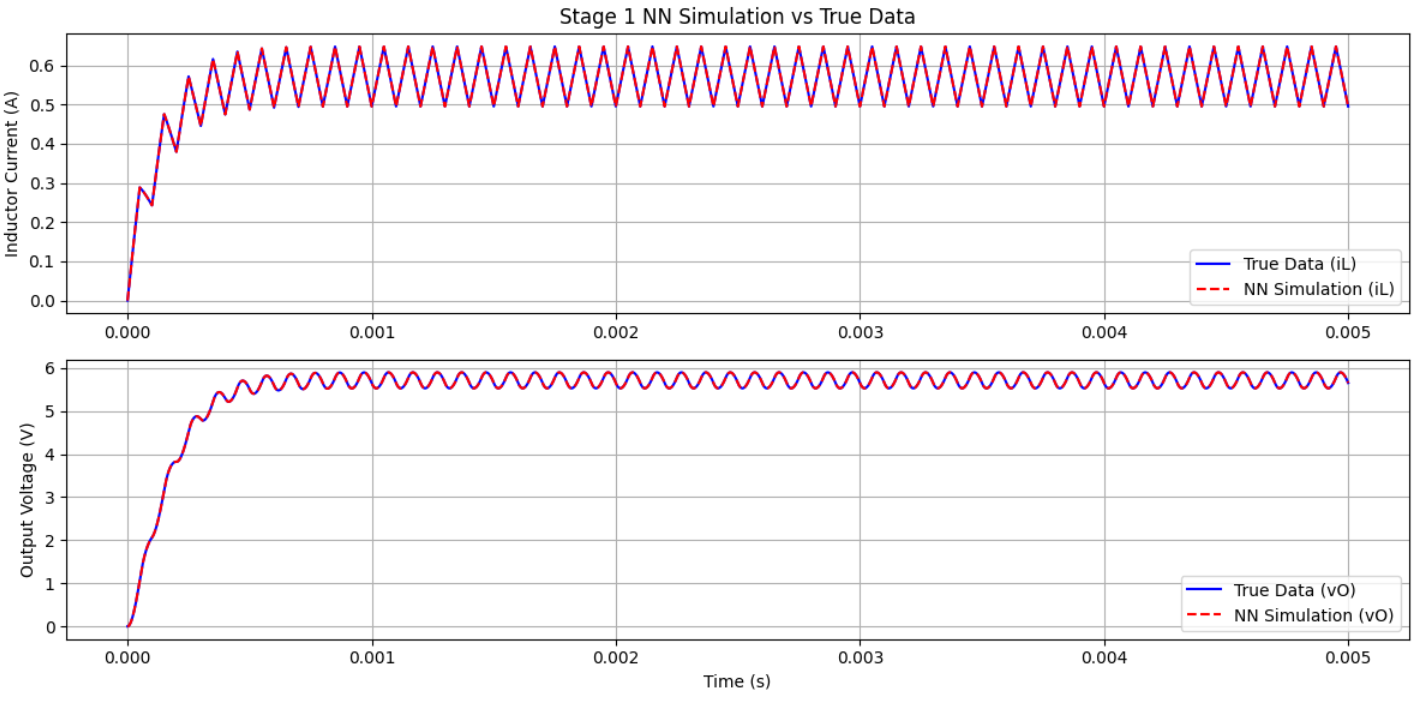}
    \caption{Stage 1 - True Data vs NN Simulation Plots}
    \label{fig:Stag1_True_vs_NN}
\end{figure*}

The input parameters for the NN model is initial values for inductor current, $i_{L0}$, the output voltage $v_{L0}$ and the Duty Cycle, \(S(t)\in\{0,1\}\). Given in Table \ref{tab:param_true_value} are the true values used during the model training. Figure \ref{fig:Stag1_True_vs_NN} shows the neural-network trajectory (dashed red) overlaid on the measured inductor current and capacitor-voltage traces (solid blue) from the synthetic dataset. Because the network has only experienced the mean-squared data loss of Stage 1, the close agreement in this panel demonstrates that the learnt derivatives integrate forward in time with no substantial drift, even across the whole 5 ms window.

\begin{table}[b]
  \centering
  \caption{Sequential-training stages required for each identification scenario}
  \label{tab:scenarios}
  \small
  \resizebox{=0.9\columnwidth}{!}{%
  \begin{tabular}{@{}p{0.65\columnwidth}c@{}}
    \toprule
    \textbf{Scenario} & \textbf{Stages of learning} \\ \midrule
    Parasitic inductor resistance ($r_L$ only)  & 2 \\
    Parasitic capacitor resistance ($r_C$ only) & 2 \\
    \multirow{2}{*}{Both $r_L$ and $r_C$} & 3 \\[-2pt]
                                           & 5\textsuperscript{*} \\ \bottomrule
  \end{tabular}}%
  \vspace{2pt}
  \footnotesize
\end{table}

\begin{table}[]
\centering
\caption{Buck Converter  Parameter True Values}
\label{tab:param_true_value}
\resizebox{=0.7\columnwidth}{!}{%
\begin{tabular}{ll}
\hline\hline
\textbf{Component Parameter}                   & \textbf{True Values} \\ \hline\hline
Inductor, $L$                       & 2$m$\textit{$H$}             \\
Capacitor, $C$                      & 5$\mu$\textit{$F$}             \\
Parasitic Inductor Resistance, $r_L$  & 0.5\textit{$\Omega$}                  \\ 
Parasitic Capacitor Resistance, $r_C$ & 30$k$\textit{$\Omega$}  \\\hline         
\end{tabular}%
}
\end{table}

\begin{table}[]
  \centering
  \caption{Common hyper-parameters used in both training schedules}
  \label{tab:common_hparams}
  \small
  \begin{tabular}{@{}lp{0.55\columnwidth}@{}}
    \toprule
    \textbf{Setting} & \textbf{Value / Rationale} \\
    \midrule
    Optimizer              & Adam $\bigl[\beta_1=0.9,\;\beta_2=0.999\bigr]$ \\[2pt]
    Learning rate          & $1\times10^{-3}$ during data pre-fit; $1\times10^{-4}$ for every parameter-release stage \\[2pt]
    Batch size             & Full record (5\,001 time-samples, $5\,$ms, $\Delta t=1\,\upmu$s) \\[2pt]
    Gradient clip          & $\lVert\nabla\theta\rVert_2\le1.0$ to avoid spikes when a parameter is unfrozen \\[2pt]
    Input scaling          & $z$-score on $i_L$, $v_O$, and duty ratio $S$ (mean 0, std 1) \\[2pt]
    Target scaling         & Same $z$-score applied to $\dot{i}_L$ and $\dot{v}_O$; data-loss weights $w_i=w_v=1$ \\[2pt]
    Physics residual scale & Residuals divided by $\sigma_{\dot{i}_L}$, $\sigma_{\dot{v}_O}$; physics weight $\lambda_{\text{phys}}=1$ \\[2pt]
    Weight initialization  & He-uniform for NN layers; log-parameters for $L$, $C$, $R_L$, $R_C$ set to nominal guesses \\[2pt]
    Random seed            & Fixed to make three- and five-stage scenarios run directly comparable \\
    \bottomrule
  \end{tabular}
\end{table}



\begin{figure*}[tbp]
    \centering
    \includegraphics[width=0.8\textwidth]{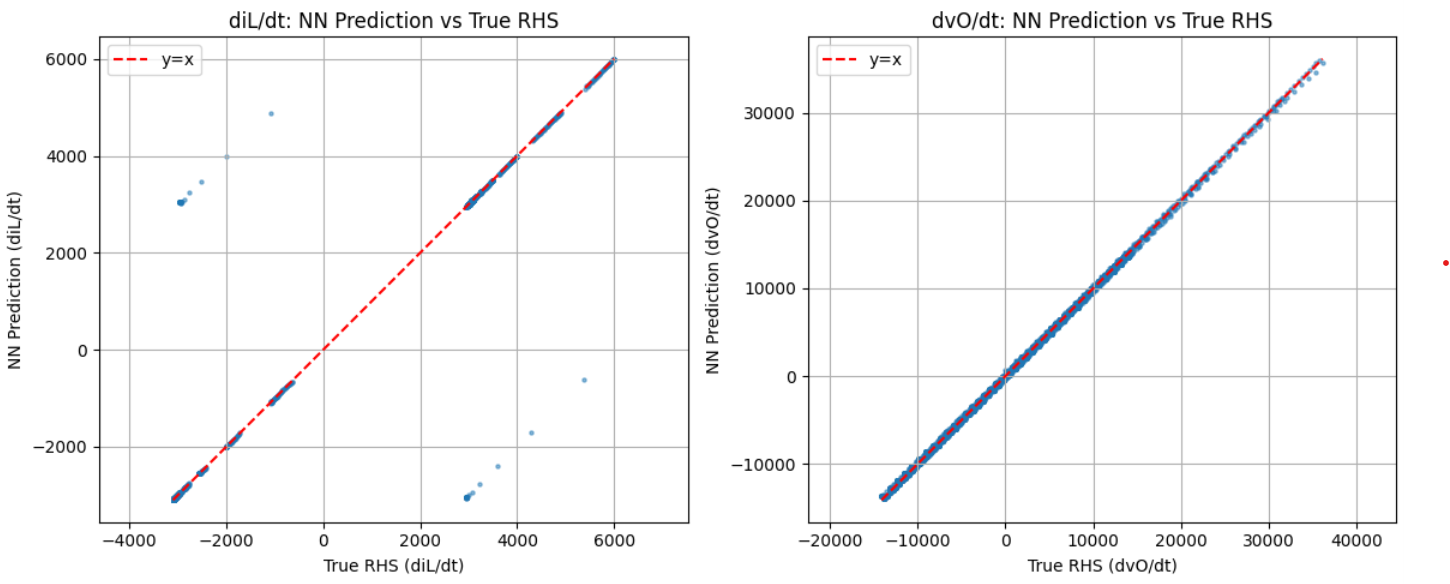}
    \caption{Stage 1 - True Data vs NN Simulation Plots}
    \label{fig:Stag1_deriv}
\end{figure*}

Figure \ref{fig:Stag1_deriv} shows a complementary point-wise examination. Each scatter plot compares the network's instantaneous derivative prediction to the analytic right-hand side based on the true component values; the 45° reference line indicates the best-fit line for perfect assimilation between the two data-points. The strong clustering around this line implies that the Stage 1 model well reflects the local slopes of both state variables. Together, the two figures show that the solely data-driven surrogate accurately reproduces the converter's dynamics, establishing a solid baseline before physics penalties are included in further training rounds.

Table \ref{tab:common_hparams} shows the training hyperparameter used to realize the models used in this work. As shown, the random seed for the training are kept the same for the two scenarios considered to maintain comparability. The Adam optimizer is used to optimize the loss function for the model with gradient clipping enabled to avoid spike during gradient learning at every stage considered.

\vspace{3 pt}\section{Results and Discussion} \label{sec:ResultsandDiscussion} \vspace{8pt}

In this section, the training schedules and the parameter sensitivity sweep of component uncertainty for the trained model are provided to demonstrate the efficacy of the staged-approach as described earlier.

\begin{table}[!h]
  \centering
  \caption{Parameter-sweep grid used for model-robustness analysis}
  \label{tab:sweep}
  \small
  \begin{tabular}{c@{\hspace{1em}}ccccc}
    \toprule
    \bfseries Offset (\%) &
    \(L\) (\(\mathrm{mH}\)) & \(C\) (\(\mu\mathrm{F}\)) & \(r_L\) (\(\Omega\)) & \(r_C\) (\(\mathrm{k}\Omega\)) \\
    \midrule
     --8 & 1.84 & 4.60 & 0.460 & 27.6 \\
     --6 & 1.88 & 4.70 & 0.470 & 28.2 \\
     --4 & 1.92 & 4.80 & 0.480 & 28.8 \\
     --2 & 1.96 & 4.90 & 0.490 & 29.4 \\
     +2  & 2.04 & 5.10 & 0.510 & 30.6 \\
     +4  & 2.08 & 5.20 & 0.520 & 31.2 \\
     +6  & 2.12 & 5.30 & 0.530 & 31.8 \\
     +8  & 2.16 & 5.40 & 0.540 & 32.4 \\
    \bottomrule
  \end{tabular}
  \vspace{2pt}
  \footnotesize
  \vspace{5pt}\\
\end{table}  

\subsection{Training Schedules} \vspace{3pt}

\begin{figure*}[t]
  \centering
  \begin{minipage}[t]{0.48\textwidth}
    \centering
    \includegraphics[width=\linewidth]{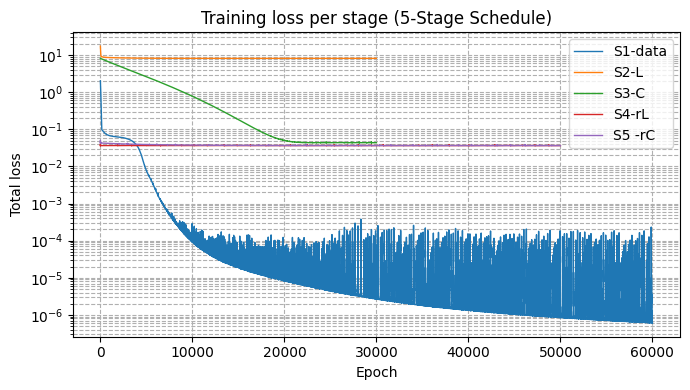}
    \caption{Training loss per stage (five-stage schedule).}
    \label{fig:TL_5stg}
  \end{minipage}\hfill
  \begin{minipage}[t]{0.48\textwidth}
    \centering
    \includegraphics[width=\linewidth]{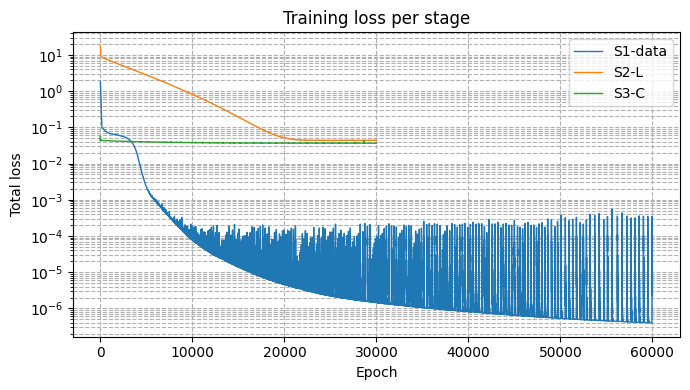}
    \caption{Training loss per stage (three-stage schedule).}
    \label{fig:TL_3stg}
  \end{minipage}
\end{figure*}

\begin{figure*}[tbp]
    \centering
    \includegraphics[width=\textwidth]{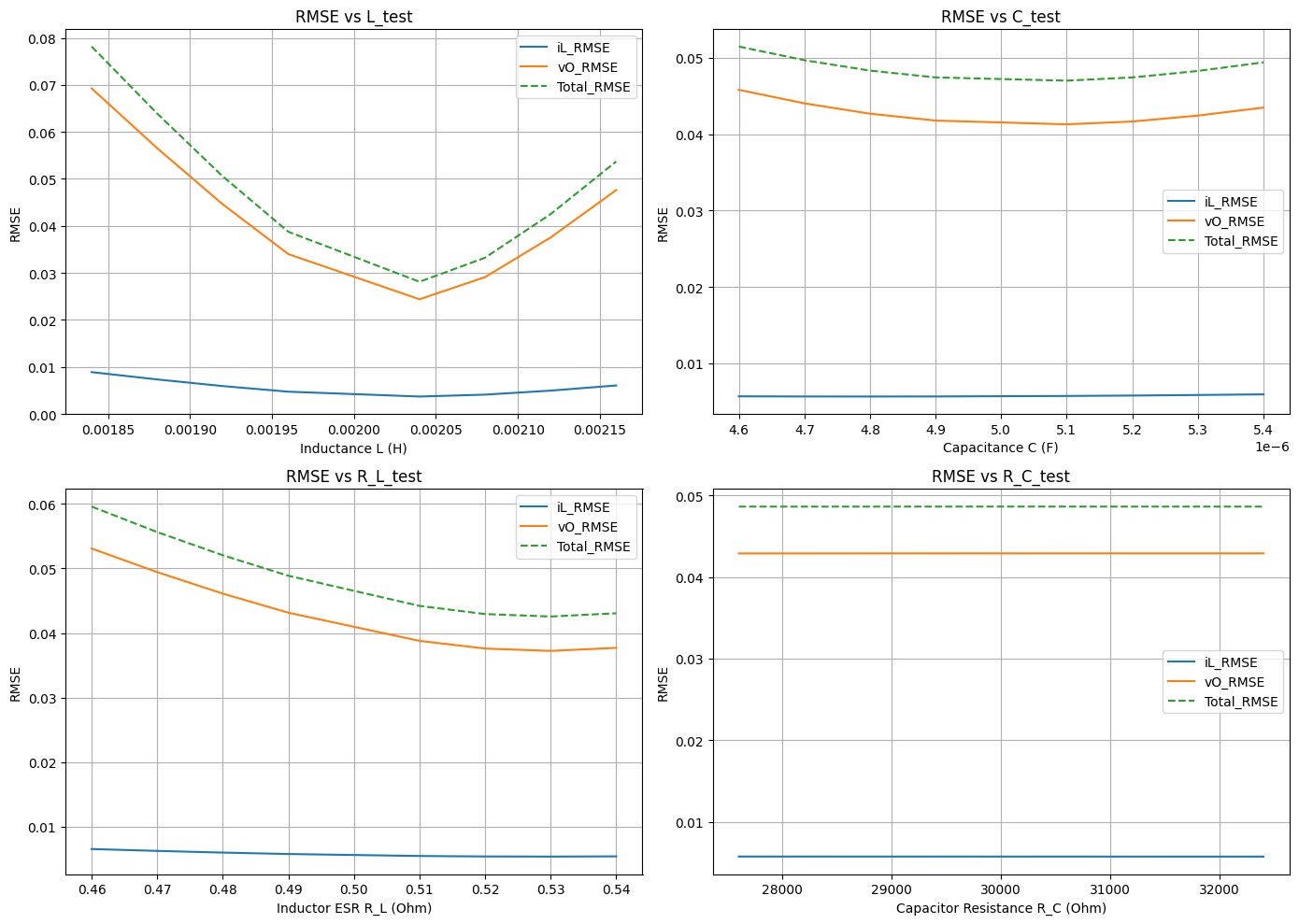}
    \caption{RMSE Plots for Parameter Sweep Grid }
    \label{fig:sweep_rmse}
\end{figure*}

\begin{table*}[t]
  \centering
  \begin{minipage}{\textwidth}
    \centering
    \subcaption{Five-Stage Training Scenario}
    \label{tab:5stage}
    \footnotesize
    \begin{tabular}{lccccccc}
      \toprule
      \textbf{Stage} & \textbf{Epochs} &
      $L$ ($\mathrm{mH}$) & $C$ ($\mu\mathrm{F}$) & 
      $r_L$ ($\Omega$) & $r_C$ ($\mathrm{k}\Omega$) &
      \textbf{Err.\ $L,C$} & \textbf{Err.\ $r_L,r_C$} \\ \midrule
      1 (data)   & 20k & \text{1.00} & \text{1.00} & 0.10 & 10 & –50\,\%, –80\,\% & –80\,\%, –67\,\% \\
      2 (+$L$)   & 50k & \textbf{2.08} & \text{1.00} & 0.10 & 10 & +4\,\%, –80\,\%  & –80\,\%, –67\,\% \\
      3 (+$C$)   & 50k & 2.08 & \textbf{5.00024} & 0.10 & 10 & +4\,\%, +0.05\,\% & –80\,\%, –67\,\% \\
      4 (+$r_L$) & 50k & 2.08 & 5.00 & \textbf{0.552} & 10 & +4\,\%, +0.05\,\% & +10\,\%, –67\,\% \\
      5 (+$r_C$) & 50k & 2.08 & 5.00 & 0.552 & \textbf{24} & +4\,\%, +0.05\,\% & +10\,\%, –20\,\% \\ \bottomrule
    \end{tabular}
  \end{minipage}

  \vspace{8pt} 

  \begin{minipage}{\textwidth}
    \centering
    \subcaption{Three-Stage Training Scenario}
    \label{tab:3stage}
    \footnotesize
    \begin{tabular}{lccccccc}
      \toprule
      \textbf{Stage} & \textbf{Epochs} &
      $L$ ($\mathrm{mH}$) & $C$ ($\mu\mathrm{F}$) & $r_L$ ($\Omega$) & $r_C$ ($\mathrm{k}\Omega$) &
      \textbf{Err.\ $L,C$} & \textbf{Err.\ $r_L,r_C$} \\ \midrule
      1 (data)          & 60k & \text{1.00} & \text{1.00} & 0.10 & 10 & –50\,\%, –80\,\% & –80\,\%, –67\,\% \\
      2 (+$L,C$)        & 50k  & \textbf{2.08} & \textbf{5.00024} & 0.10 & 10 & +4\,\%, +0.05\,\%    & –80\,\%, –67\,\% \\
      3 (+$r_L,r_C$)    & 45k  & 2.08 & 5.00 & \textbf{0.552} & \textbf{24} & +4\,\%, +0.05\,\% & +10\,\%, –20\,\% \\ \bottomrule
    \end{tabular}
  \end{minipage}
\end{table*}





Tables~\ref{tab:5stage} and~\ref{tab:3stage} describe the parameter trajectories for the five-stage and three-stage sequential-PINN schedules, respectively. During the first stage of the experiment, the network is trained using the training data to provide a universal approximation for the overall system based on the ODE. In the next stages of learning, the inductance and capacitance are retained at their inferior initial estimates (\(\approx -50\%\) and \(\approx -80\%\), respectively), while both parasitic resistive parameters are similarly set as untrained estimate values. Therefore, freezing the PDE gradients from the training procedure.

 In the five-stage training schedule case, the inductance estimate is reduced to \(2.08~\text{$mH$}\) (\(+4\%\) error) by releasing the inductance learning gradient in Stage 2 for \(50\,\textit{k}\) training epochs, with no effect on the other values. In Stage 3, the capacitance learning gradient is unfrozen and exactly converges to the true value of \(5~\mu\text{$F$}\) with an error of \(240~p\text{$F$}\)(\(+0.05\%\)), leaving the previously estimated inductance unaltered. The parasitic resistance for inductance  and capacitance learning gradient is allowed in Stages 4 and 5, respectively. In Stage 4, the series coil resistance is modified to \(0.552~\Omega\) (\(\approx +10\%\)), and the parallel leakage resistance, \(r_{C}\)  pathway is released for the last training stage. This resulted in a reduction in the \(r_{C}\) error from \(-67\%\) to \(-20\%\) demonstrating that the gradual release of gradients keeps earlier parameters from drifting. 
 
The three-stage training schedule compresses the technique, combining \(L\) and \(C\) into a single learning phase (stage 2) and the two resistive parameters into a second (Stage 3), maintaining the same training strategy at Stage 1 as above. During Stage 2, in \(50,\textit{k}\) epochs, the inductance is decreased to \(2.08~\text{$mH$}\) and the capacitance to \(5.00024~\mu\text{$F$}\) by the joint release of \(L\) and \(C\) following a \(50\,\text{k}\)-epoch data-only pre-fit in Stage 1. This implies that fine-grained staging isn't strictly required for convergence.

\begin{figure*}[!t]
  \centering
  \includegraphics[width=\textwidth]{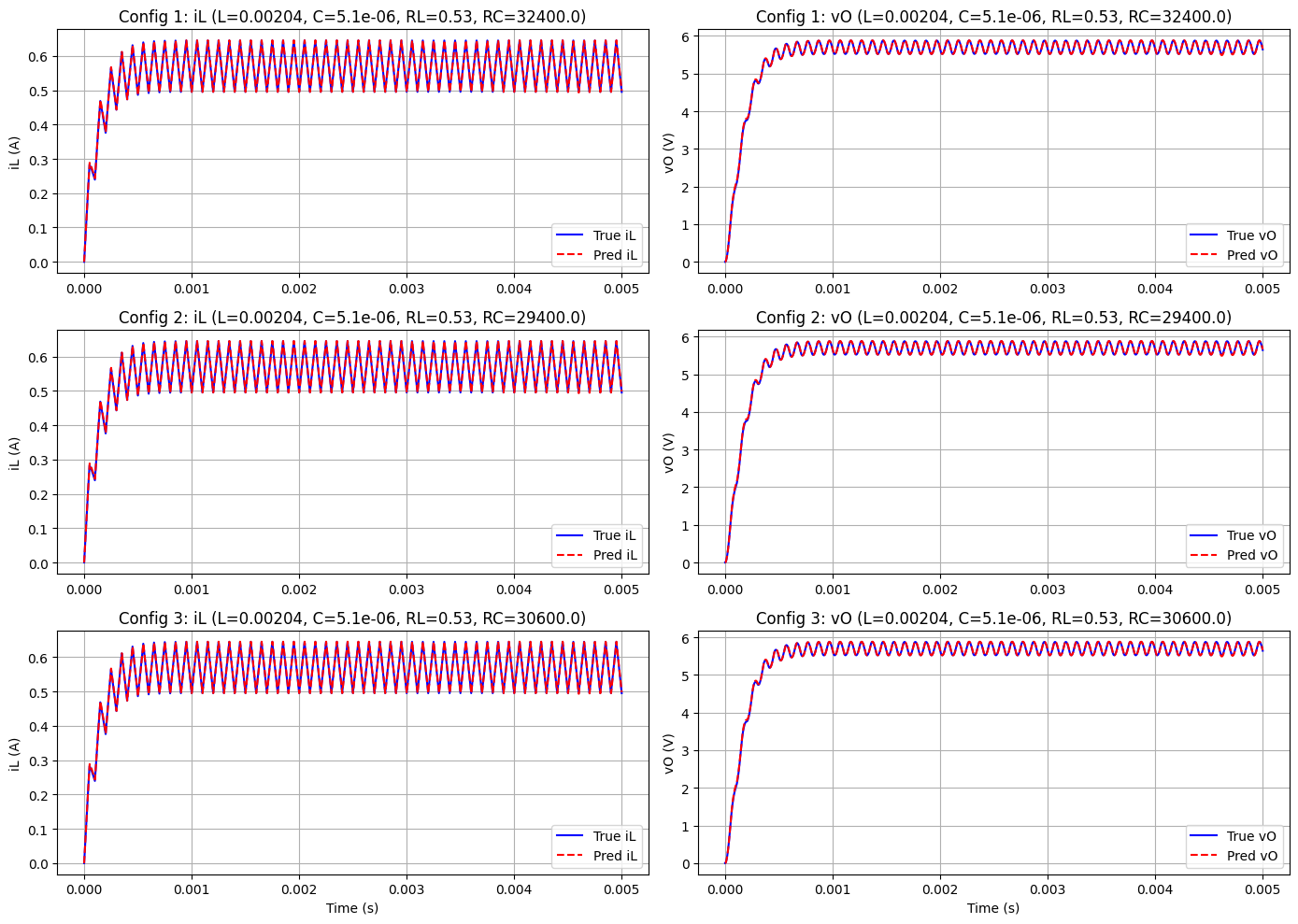}
  \caption{PINN (red) versus truth (blue) for the three best parameter sets.}

  \label{fig:top3_time}
\end{figure*}

Figures \ref{fig:TL_5stg} and \ref{fig:TL_3stg} compare the convergence behaviour of the five-stage and three-stage training schedules.
In both cases the log-scaled loss drops steeply during the data-only pre-fit, indicating that the neural network alone can approximate the measured derivatives with little optimization effort.
For the five-stage run the loss then decreases in a series of smaller, step-like reductions as \(L\), \(C\), \(R_{L}\) and \(R_{C}\) are
sequentially unfrozen; each parameter release introduces a brief spike, followed by rapid settling to a lower plateau.

By contrast, the three-stage curve shows only two such plateaux, because \(L\) and \(C\) are learned together in Stage 2, and the two parasitics are learned together in Stage 3. Although the three-stage loss reaches its final plateau about 10\,\% sooner in time, the five-stage trace reveals a more gradual, monotonic descent, suggesting a smoother optimization path and better interpretability of intermediate parameter values.

The residual error is essentially limited to the parallel leakage resistance in both schedules. The high impedance of \(r_{C}\) creates a weak signature in the state derivatives, resulting in a tiny gradient that is easily disguised by numerical noise. The remaining \(20\%\) uncertainty in \(r_{C}\) is unimportant for loop-shaping and transient-response investigations, while the accurate recovery of \(L\), \(C\), and \(r_{L}\) is enough for extensive efficiency and control analysis. As a result, a viable workflow begins with the compact three-stage schedule to obtain reliable core parameters and advances to the five-stage sequence only when a higher leakage resistance boundary is required.


Table~\ref{tab:sweep} depicts the eight-point grid used in the sensitivity sweep. To achieve offsets of \(\pm 2\%,\;\pm 4\%,\;\pm 6\%\) and \(\pm 8\%\), each component value is symmetrically perturbed about its nominal setting in uniform \(2~\%\) increments.
The test points' inductance vector spans \(1.84\text{-}2.16~\text{$mH$}\), capacitance vector spans \(4.60\text{-}5.40~\mu\text{$F$}\), series coil resistance vector spans \(0.460\text{-}0.540~\Omega\), and parallel leakage resistance vector spans \(27.6\text{-}32.4~\text{$k$}\Omega\).
The study assesses \(8^{4}=4096\) different parameter choices due to the factorial sweep, which is adequate to demonstrate second-order sensitivity effects without incurring high processing costs.

\subsection{Sensitivity of the trained PINN to component uncertainty}


As show in in Fig.~\ref{fig:sweep_rmse} blue, orange, and green curves correspond to the current channel ($i_{L}$-RMSE), the voltage channel ($v_{O}$-RMSE), and their sum ($\mathrm{Total\_RMSE}$), respectively. 
It summarizes the $8^{4}$-point sensitivity experiment by combining each three-parameter sub-grid into a single root-mean-square error (RMSE) trace for the remaining parameter.\vspace{5pt}

\textbf{Inductance sweep.} 
Adjusting $L$ between $1.84\! - \!2.16\,\mathrm{mH}$ results in a unique bowl shape of the total RMSE. It reaches a minimum of $2.6\times10^{-2}$ at the true value of $L = 2.00\,\mathrm{mH}$ and quickly increases to the $\pm8\%$ extremes.
The curve is dominated by the voltage component, and the current error is below $8\times10^{-3}$, indicating that the network's voltage prediction is particularly susceptible to inductance bias.\vspace{5pt}



\textbf{Capacitance sweep.} Varying $C$ between $4.60$–$5.40\,\upmu\mathrm{F}$ induces a stable, nearly parabolic trend in total RMSE. Throughout the range, the RMSE fluctuation is less than $5 \times 10^{-3}$, demonstrating remarkable stability. This confirms that increasing precision leads to declining gains in prediction fidelity after $C$ is predicted within a few percent.\vspace{5pt}

\textbf{Series ESR sweep.}  
The total error for $r_{L}$ drops monotonically in the range of $0.460\!-\!0.540\,\Omega$, reaching a minimum at $r_{L}=0.52\,\Omega$. Overestimating significantly dampens the converter and improves the fit, whereas underestimating can inflate the RMSE by about 40\%.\vspace{5pt}

\textbf{Leakage resistance sweep.} Changing $r_{C}$ between $27.6$ and $32.4\,\mathrm{k}\Omega$ leaves all traces essentially flat; a full $\pm8\%$ perturbation moves the total RMSE by less than 1 \%. The leakage path carries micro-ampere currents under normal load, because its exact value has negligible impact on one-step prediction.\vspace{5pt}



The suggested sequential-learning framework's parameter release order is supported by the sensitivity data in Fig.~\ref{fig:sweep_rmse}. The monotonic fall in the series ESR and the prominent U-shape obtained when scanning the inductance illustrate that a tiny bias in either \(L\) or \(r_{L}\) directly generates a large voltage-channel RMSE. 
The capacitance curve is nearly flat, while the leakage-resistance trace is nearly horizontal. This suggests that moderate faults in \(C\) or \(r_{C}\) have minimal impact on one-step-ahead performance.


These gradient magnitudes provide insight into the empirical behavior of the training plan. The optimizer can quickly descend without disturbing the current-channel fit formed during the data-only stage by releasing \(L\), which removes the stiffest direction in the loss surface. 
Adding \(r_{L}\) improves the voltage prediction by dampening the LC pole and stabilizing the inductance estimate. The less sensitive parameters \(C\) and \(r_{C}\) are only unfrozen when the dominant quantities have stabilized. Their fine-tuning produces small but measurable improvements in steady-state accuracy.

In summary, the sweep confirms that the sequential order \(\{L\rightarrow r_{L}\rightarrow C \rightarrow r_{C}\}\) mirrors the natural hierarchy of parameter influence, thereby reducing optimization stiffness and accelerating convergence while preserving final fidelity. The PINN reproduces the full-order response of the top three buck converter parameter sweep scenarios, as shown in Fig.~\ref{fig:top3_time}, even when all four components are affected concurrently. After an initial \(0.3\)\,$ms$ rise, the predicted inductor current and output voltage for the three lowest-RMSE configurations are visually indistinguishable from the reference traces, and the steady-state ripple exhibits no phase lag. The results demonstrate that progressively determined parameters are generalizable to unseen operating points that differ by up to \(\pm8\%\) from the nominal values. The loss surface is steepest along the inductance and ESR vectors, but virtually flat along the capacitance and leakage paths, according to the sensitivity research.


\section{Conclusion} \label{sec:Concl}
\vspace{5 pt}

In this work, the investigated five-stage schedule (\(L\rightarrow r_{L}\rightarrow C\rightarrow r_{C}\)) reduced the leakage-resistance bias from -67\% to -20\%, while driving the neural network to sub-5\% error on \(L\) and \(C\), and to 10\% on \(r_{L}\). A three-stage compression produced the same precision at an 11\% faster rate, proving that steps can be merged when time and resources are limited. No grid search or human tuning is required because each block is trained with a single, unit data-to-physics weight, and the parameter traces resolve monotonically. As a result, the staged PINN outperforms a one-shot fit in terms of accuracy, drift-free parameter estimations, computation time, and hyperparameter management. Finally, the sensitivity analysis revealed that a PINN model could potentially exhibit a stable behavior in tracking parameter changes.

\end{document}